\documentclass[11pt]{article}
\usepackage{paclic33}
\usepackage{times}
\usepackage{latexsym}
\usepackage{amsmath, amssymb}
\usepackage{mathtools}
\usepackage{url}
\usepackage{multirow}
\usepackage{graphicx}

\title{On the Effectiveness of Low-Rank Matrix Factorization for \\LSTM Model Compression}

\author{Genta Indra Winata, Andrea Madotto, Jamin Shin, Elham J. Barezi, Pascale Fung  \\
  Center for Artificial Intelligence Research (CAiRE) \\
  Department of Electronic and Computer Engineering \\
  The Hong Kong University of Science and Technology, Clear Water Bay, Hong Kong \\
  {\tt \{giwinata,amadotto,jmshinaa,ejs\}@connect.ust.hk}\\ {\tt pascale@ece.ust.hk}}
\date{}

\begin{document}
\maketitle

\begin{abstract}
Despite their ubiquity in NLP tasks, Long Short-Term Memory (LSTM) networks suffer from computational inefficiencies caused by inherent unparallelizable recurrences, which further aggravates as LSTMs require more parameters for larger memory capacity. In this paper, we propose to apply low-rank matrix factorization (MF) algorithms to different recurrences in LSTMs, and explore the effectiveness on different NLP tasks and model components. We discover that additive recurrence is more important than multiplicative recurrence, and explain this by identifying meaningful correlations between matrix norms and compression performance. We compare our approach across two settings: 1) compressing core LSTM recurrences in language models, 2) compressing biLSTM layers of ELMo evaluated in three downstream NLP tasks.
\end{abstract}

\section{Introduction}
Long Short-Term Memory (LSTM) networks~\cite{hochreiter1997long,gers1999learning} have become the core of many models for tasks that require temporal dependency. They have particularly shown great improvements in many different NLP tasks, such as Language Modeling~\cite{sundermeyer2012lstm,mikolov2012statistical}, Semantic Role Labeling~\cite{he2017deep}, Named Entity Recognition~\cite{lee2017end}, Machine Translation~\cite{bahdanau2014neural}, and Question Answering~\cite{bidaf}. Recently, a bidirectional LSTM has been used to train deep contextualized Embeddings from Language Models (ELMo)~\cite{ELMo}, and has become a main component of state-of-the-art models in many downstream NLP tasks. 

However, there is an obvious drawback of scalability that accompanies these excellent performances, not only in training time but also during inference time. This shortcoming can be attributed to two factors: the temporal dependency in the computational graph, and the large number of parameters for each weight matrix. The former problem is an intrinsic nature of RNNs that arises while modeling temporal dependency, and the latter is often deemed necessary to achieve better generalizability of the model \cite{hochreiter1997long,gers1999learning}. On the other hand, despite such belief that the LSTM memory capacity is proportional to model size, several recent results have empirically proven the contrary, claiming that LSTMs are indeed over-parameterized \cite{denil2013predicting,qrnn2017,merity2018regularizing,melis2018on,lstm_weighted_sum}. 

Naturally, such results motivate us to search for the most effective compression method for LSTMs in terms of performance, time, and practicality, to cope with the aforementioned issue of scalability. There have been many solutions proposed to compress such large, over-parameterized neural networks including parameter pruning and sharing~\cite{gong2014compressing,huang2018learning}, low-rank Matrix Factorization (MF)~\cite{jaderberg2014speeding}, and knowledge distillation~\cite{hinton2015distilling}. However, most of these approaches have been applied to Feed-forward Neural Networks and Convolutional Neural Networks (CNNs), while only a small attention has been given to compressing LSTM architectures~\cite{lu2016learning,belletti2018factorized}, and even less in NLP tasks. Notably, \shortcite{see2016compression} applied parameter pruning to standard Seq2Seq~\cite{sutskever2014sequence} architecture in Neural Machine Translation, which uses LSTMs for both encoder and decoder. Furthermore, in language modeling, \shortcite{grachev2017neural} uses Tensor-Train Decomposition~\cite{oseledets2011tensor}, \shortcite{N18-1192} uses binarization techniques, and \shortcite{kuchaiev2017factorization} uses an architectural change to approximate low-rank factorization. 

All of the above mentioned works require some form of training or retraining step. For instance, \shortcite{kuchaiev2017factorization} requires to be trained completely from scratch, as well as distillation based compression techniques~\cite{hinton2015distilling}. In addition, pruning techniques~\cite{see2016compression} often accompany selective retraining steps to achieve optimal performance. However, in scenarios involving large pre-trained models, e.g., ELMo~\cite{ELMo}, retraining can be very expensive in terms of time and resources. Moreover, compression methods are normally applied to large and over-parameterized networks, but this is not necessarily the case in our paper. We consider strongly tuned and regularized state-of-the-art models in their respective tasks, which often already have very compact representations. These circumstances make the compression much more challenging, but more realistic and practically useful. 

In this work, we advocate low-rank matrix factorization as an effective post-processing compression method for LSTMs which achieve good performance with guaranteed minimum algorithmic speed compared to other existing techniques. We summarize our contributions as the following:
\begin{itemize}
    \item We thoroughly explore the limits of several different compression methods (matrix factorization and pruning), including fine-tuning after compression, in Language Modeling, Sentiment Analysis, Textual Entailment, and Question Answering.
    \item We consistently achieve an average of 1.5x (50\% faster) speedup inference time while losing $\sim$1\ point in evaluation metric across all datasets by compressing additive and/or multiplicative recurrences in the LSTM gates. 
    \item In PTB, by further fine-tuning very compressed models ($\sim$98\%) obtained with both matrix factorization and pruning, we can achieve $\sim$2x (200\% faster) speedup inference time while even slightly improving the performance of the uncompressed baseline.
    \item We discover that matrix factorization performs better in general, additive recurrence is often more important than multiplicative recurrence, and we identify clear and interesting correlations between matrix norms and compression performance.
\end{itemize}

\section{Related Work}
The current approaches of model compression are mainly focused on matrix factorization, pruning, and quantization. The effectiveness of these approaches were shown and applied in different modalities. In speech processing,~\shortcite{wilson2008speech,mohammadiha2013supervised,geiger2014investigating,fan2014speech} studied the effectiveness of Non-Matrix Factorization (NMF) on speech enhancement by reducing the noisy speech interference. Matrix factorization-based techniques were also applied in image captioning \cite{hong2016joint,li2017graph} by exploiting the clustering intepretations of NMF. Semi-NMF, proposed by~\shortcite{ding2010convex}, relaxed the constraints of NMF to allow mixed signs and extend the possibility to be applied in non-negative cases.~\shortcite{trigeorgis2014deep} proposed a variant of the Semi-NMF to learn low-dimensional representation through a multi-layer structure.~\shortcite{W18-3410} proposed to replace GRUs with low-rank and diagonal weights to enable low-rank parameterization of LSTMs. \shortcite{kuchaiev2017factorization} modifed LSTM structure by replacing input and hidden weights with two smaller partitions to boost the training and inference time.

% Furthermore, many matrix factorization methods have been extensively developed to reduce the redundancy on over-parameterized deep neural networks as low-rank minimization problem~\cite{denil2013predicting,qrnn2017,merity2018regularizing,melis2018on}. 
% These methods were applied in the structure of RNNs for obtaining low-rank approximated representations.
On the other hand, compression techniques can also be applied as post-processing steps.~\shortcite{grachev2017neural} investigated low-rank factorization on standard LSTM model. The Tensor-Train method has been used to train end-to-end high-dimensional sequential video data with LSTM and GRU \cite{yang2017tensor,tjandra2017compressing}. In another line of work, \shortcite{K16-1029} explored pruning in order to reduce the number of parameters in Neural Machine Translation. \shortcite{wen2018learning} proposed to zero out the weights in the network learning blocks to remove insignificant weights of the RNN. Meanwhile, \shortcite{N18-1192} proposed to binarize LSTM Language Models. Finally, \shortcite{han2016deep} proposed to use all pruning, quantization, and Huffman coding to the weights on AlexNet. 

\section{Methodology}

\begin{figure*}[t]
    \centering
    \includegraphics[width=0.82\linewidth]{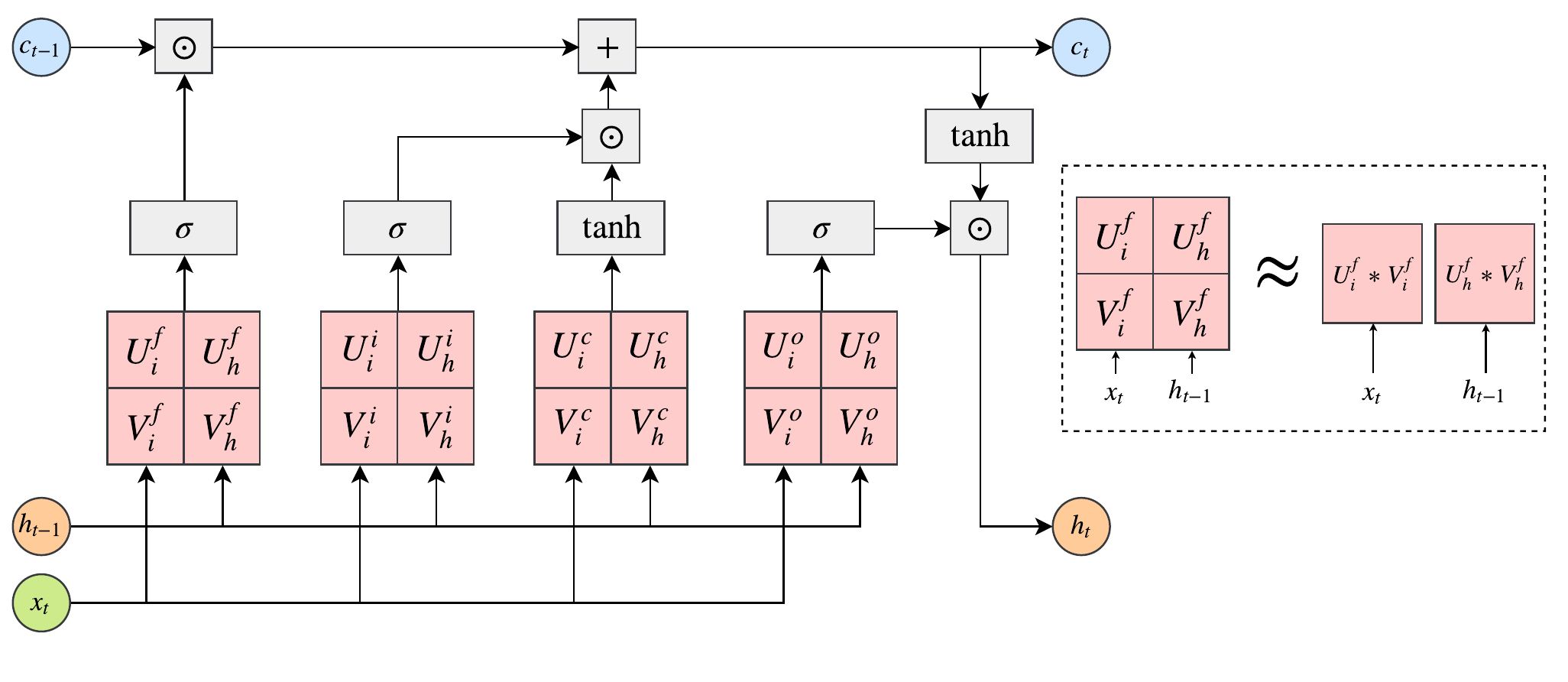}
    \caption{Factorized LSTM Cell}
    \label{fig:lSTM}
\end{figure*}

% \begin{figure*}[ht!]
%     \centering
%     \includegraphics[scale=0.35]{images/diagram.pdf}
%     \caption{Factorized LSTM Cell}
%     \label{fig:lSTM}
% \end{figure*}

\subsection{Long-Short Term Memory Networks}
Long-Short Term Memory (LSTMs) networks are parameterized with two large matrices, $\mathbf{W}_i$, and $\mathbf{W}_h$. LSTM captures long-term dependencies in the input and avoids the exploding/vanishing gradient problems on the standard RNN. The gating layers control the information flow within the network and decide which information to keep, discard, or update in the memory. The following recurrent equations show the LSTM dynamics:
\begin{align}
    \small
    \label{eq:LSTM_gate}
    \begin{pmatrix} 
        \mathbf{i}_t \\ 
        \mathbf{f}_t \\ 
        \mathbf{o}_t \\ 
        \mathbf{\hat{c}}_t 
    \end{pmatrix}
    = 
    \begin{pmatrix} 
        \sigma \\ 
        \sigma \\ 
        \sigma \\ 
        \text{tanh}
    \end{pmatrix}
    % \odot
    \begin{pmatrix} 
        \mathbf{W}_i & \mathbf{W}_h
    \end{pmatrix}
    \begin{pmatrix} 
        \mathbf{x}_t \\
        \mathbf{h}_{t-1}
    \end{pmatrix},
\\
    \small
    \mathbf{W}_i = 
    \begin{pmatrix} 
        \mathbf{W}_i^i \\ 
        \mathbf{W}_i^f \\ 
        \mathbf{W}_i^o \\ 
        \mathbf{W}_i^c 
    \end{pmatrix},
    \mathbf{W}_h = 
    \begin{pmatrix} 
        \mathbf{W}_h^i \\ 
        \mathbf{W}_h^f \\ 
        \mathbf{W}_h^o \\ 
        \mathbf{W}_h^c 
    \end{pmatrix},
\end{align}
\begin{equation}\label{eq:LSTM}
\begin{aligned} 
\mathbf{c}_t =&~ \mathbf{f}_t \odot \mathbf{c}_{t-1} + \mathbf{i}_t  \odot \mathbf{\hat c}_t, \\ 
\mathbf{h}_t =&~ \mathbf{o}_t \odot \text{tanh}(\mathbf{c}_t).
\end{aligned}
\end{equation}
where $\mathbf{x}_t \in \mathbb{R}^{n_{inp}}$,  and $\mathbf{h}_t \in \mathbb{R}^{n_{dim}}$ at time $t$. Here, $\sigma(\cdot)$ and $\odot$ denote the sigmoid function and element-wise multiplication operator, respectively. The model parameters can be summarized in a compact form with: $\Theta = [\mathbf{W}_i, \mathbf{W}_h]$, where $\mathbf{W}_{i}\in \mathbb{R}^{4*n_{inp} \times 4*n_{dim} }$ which is the input matrix, and $\mathbf{W}_{h} \in \mathbb{R}^{4*n_{dim} \times 4*n_{dim} }$ which is the hidden matrix. Note that we often refer $\mathbf{W}_i$ as additive recurrence and $\mathbf{W}_h$ as multiplicative recurrence, following terminology of \shortcite{lstm_weighted_sum}.

\subsection{Low-Rank Matrix Factorization}
We consider two Low-Rank Matrix Factorization for LSTM compression: Truncated Singular Value Decomposition (SVD) and Semi Non-negative Matrix Factorization (Semi-NMF). Both methods factorize a matrix $\mathbf{W}$ into two matrices $\mathbf{U}_{m\times r}$ and $\mathbf{V}_{r\times n}$ such that $\mathbf{W} = \mathbf{U}\mathbf{V}$ \cite{fazel2002matrix}. SVD produces a factorization by applying orthogonal constraints on the $\mathbf{U}$ and $\mathbf{V}$ factors along with an additional diagonal matrix of singular values, where instead Semi-NMF generalizes Non-negative Matrix Factorization (NMF) by relaxing some of the sign constraints on negative values for $\mathbf{U}$ and $\mathbf{W}$.
The computation advantage, compared to pruning methods which require a special implementation of sparse matrix multiplication, is that the matrix $\boldsymbol{\mathbf{W}}$ requires $mn$ parameters and $mn$ flops, while $\mathbf{U}$ and $\mathbf{V}$ require $rm + rn=r(m+n)$ parameters and $r(m+n)$ flops. If we take the rank to be very low $r<<m,n$, the number of parameters in $\mathbf{U}$ and $\mathbf{V}$ is much smaller compared to $\mathbf{W}$. 

As elaborated in Equation \ref{eq:LSTM_gate}, a basic LSTM cell includes four gates: input, forget, output, and cell state, performing a linear combination on input at time $t$ and hidden state at time $t-1$. We propose to replace $\mathbf{W}_i, \, \mathbf{W}_h$ pair for each gate with their low-rank decomposition, either SVD or Semi-NMF \cite{ding2010convex}, leading to a significant reduction in memory and computational cost requirement. The general objective function is given as:
\begin{align}
\label{eq:factorization}
\underset{m \times n}{\mathbf{W}} = \underset{m \times r} {\mathbf{U}} \quad \underset{r \times n}{\mathbf{V}}, \\
\underset{U, V}{\text{minimize}}
\quad ||\mathbf{W} - \mathbf{U} \mathbf{V}||_F^2.
\end{align}

\subsection{Truncated Singular Value Decomposition (SVD)}
One of the constrained matrix factorization method is based on Singular Value Decomposition (SVD) which produces a factorization by applying orthogonal constraints on the $\mathbf{U}$ and $\mathbf{V}$ factors. These approaches aim to find a linear combination of the basis vectors which restrict to the orthogonal vectors in feature space that minimize reconstruction error.
In the case of the SVD, there are no restrictions on the signs of $\mathbf{U}$ and $\mathbf{V}$ factors. Moreover, the data matrix $\mathbf{W}$ is also unconstrained. 

\begin{align}
\label{eq:nmf0}
\mathbf{W} = \mathbf{U} \mathbf{S} \mathbf{V}, \\ \underset{\mathbf{U}, \mathbf{S}, \mathbf{V}}{\text{minimize}}
\quad ||\mathbf{W} - \mathbf{U} \mathbf{S} \mathbf{V}||_F^2.
\end{align}

s.t. $\mathbf{U}$ and $\mathbf{V}$ are orthogonal, and $\mathbf{S}$ is diagonal. The optimal values $\mathbf{U}_{m \times r}^r$, $\mathbf{S}_{r \times r}^r$, $\mathbf{V}_{r \times n}^r$ for $\mathbf{U}_{m\times n}$, $\mathbf{S}_{n\times n}$, and $\mathbf{V}_{n\times n}$ are obtained by taking the top $r$ singular values from the diagonal matrix $\mathbf{S}$ and the corresponding singular vectors from $\mathbf{U}$ and $\mathbf{V}$.

\subsection{Semi-NMF}
Semi-NMF generalizes Non-negative Matrix Factorization (NMF) by relaxing some of the sign constraints on negative values for $\mathbf{U}$ and $\mathbf{W}$ ($\mathbf{V}$ has to be kept positive). Semi-NMF is more preferable in application to Neural Networks because of this generic capability of having negative values. To elaborate, when the input matrix $\mathbf{W}$ is unconstrained (i.e., contains mixed signs), we consider a factorization, in which we restrict $\mathbf{V}$ to be non-negative, while having no restriction on the signs of $\mathbf{U}$. We minimize the objective function as in Equation \ref{eq:semi}.

\begin{align}
\label{eq:semi}
\mathbf{W}_{\pm} \approx \mathbf{U}_{\pm} \mathbf{V}_+, \\
\underset{\mathbf{U}, \mathbf{V}}{\text{minimize}}
\quad ||\mathbf{W} - \mathbf{U} \mathbf{V}||_F^2 \quad \text{s.t. } \mathbf{V}\geq 0. 
\end{align}

The optimization algorithm iteratively alternates between the update of $\mathbf{U}$ and $\mathbf{V}$ using coordinate descent~\cite{luo1992convergence}.

\subsection{Pruning}
We use the pruning methodology used in LSTMs from~\shortcite{han2015} and~\shortcite{K16-1029}. To elaborate, for each weight matrix $\mathbf{W}_{i,h}$, we mask the low-magnitude weights to zero, according to the compression ratio of the low-rank factorization\footnote{We align the pruning rate with the rank with $\frac{r(m+n)}{mn}$.}.

\begin{table}[t]
\centering
\caption{The table shows the total parameters, perplexity, and compression efficiency (lower is better) on PTB Language Modeling task. $\ddagger$ We reproduced the results.}
\resizebox{0.5\textwidth}{!}{
\begin{tabular}{lccccc}
\hline
\multicolumn{1}{c}{\multirow{2}{*}{\textbf{\textbf{PTB}}}} & \multirow{2}{*}{\textbf{Param.}} & \multicolumn{2}{c}{\textbf{w/o fine-tuning}} & \multicolumn{2}{c}{\textbf{w/ fine-tuning}} \\ \cline{3-6} 
 & & \textbf{PPL} & \textbf{E(r)} & \textbf{PPL} & \textbf{E(r)} \\ \hline \hline
AWD-LSTM & 24M & 58.3{$^\ddagger$} & - & 57.3 & - \\
TT-LSTM & 12M & 168.6 & 2.92{$^\dagger$} & - & - \\ \hline \hline
Semi-NMF $\mathbf{W}_h$ \small{(r=10)} & 9M & 78.5 & 0.72 & 58.11 & -0.02 \\
SVD $\mathbf{W}_h$ \small{(r=10)} & 9M & 78.07 & 0.32 & 58.18 & -0.02 \\
Pruning $\mathbf{W}_h$ \small{(r=10)} & 9M & 83.62 & 0.89 & 57.94 & -0.03 \\ \hline \hline
Semi-NMF $\mathbf{W}_h$ \small{(r=400)} & 18M & 59.7 & 0.05 & 57.84 & -0.02 \\
SVD $\mathbf{W}_h$ \small{(r=400)} & 18M & \textbf{59.34} & \textbf{0.006} & 57.81 & -0.02 \\
Pruning $\mathbf{W}_h$ \small{(r=400)} & 18M & 59.47 & 0.03 &  \textbf{57.19} & \textbf{-0.04} \\ \hline \hline
Semi-NMF $\mathbf{W}_i$ \small{(r=10)} & 15M & 485.4 & 19.81 & 81.4 & 1.04 \\
SVD $\mathbf{W}_i$ \small{(r=10)} & 15M & 462.19 & 6.83 & 88.12 & 1.35 \\
Pruning $\mathbf{W}_i$ \small{(r=10)} & 15M & 676.76 & 28.69 & 82.23 & 1.08 \\ \hline \hline
Semi-NMF $\mathbf{W}_i$ \small{(r=400)} & 20M & 62.7 & 0.42 & 58.47 & -0.01 \\
SVD $\mathbf{W}_i$ \small{(r=400)} & 20M & 60.59 & 0.02 & 58.04 & -0.01 \\
Pruning $\mathbf{W}_i$ \small{(r=400)} & 20M & 59.62 & 0.10 & 57.65 & -0.02 \\ \hline
\end{tabular}
}
\label{lm}
\end{table}

\begin{table}[t]
\centering
\caption{The table shows the total parameters, perplexity, and compression efficiency (lower is better) on WT-2 Language Modeling task. $\ddagger$ We reproduced the results.}
\resizebox{0.49\textwidth}{!}{
\begin{tabular}{lccc}
\hline
\multicolumn{1}{c}{\textbf{WT-2}} & \textbf{Params} & \textbf{PPL} & \textbf{E(r)} \\ \hline \hline
AWD-LSTM & 24M & 65.67{$^\ddagger$} & - \\ \hline
Semi-NMF $\mathbf{W}_h$ (r=10) & 9M & 102.17 & 65.14 \\
SVD $\mathbf{W}_h$ (r=10) & 9M & 99.92 & 62.49 \\
Pruning $\mathbf{W}_h$ (r=10) & 9M & 109.16 & 72.64 \\ \hline \hline
Semi-NMF $\mathbf{W}_h$ (r=400) & 18M & 66.5 & 4.33 \\
SVD $\mathbf{W}_h$ (r=400) & 18M & \textbf{66.1} & \textbf{2.28} \\
Pruning $\mathbf{W}_h$ (r=400) & 18M & 66.23 & 2.94 \\ \hline \hline
Semi-NMF $\mathbf{W}_i$ (r=10) & 15M & 481.61 & 197.57 \\
SVD $\mathbf{W}_i$ (r=10) & 15M & 443.49 & 194.89 \\
Pruning $\mathbf{W}_i$ (r=10) & 15M & 856.87 & 211.23 \\ \hline \hline
Semi-NMF $\mathbf{W}_i$ (r=400) & 20M & 68.41 & 22.68 \\
SVD $\mathbf{W}_i$ (r=400) & 20M & 67.11 & 12.18 \\
Pruning $\mathbf{W}_i$ (r=400) & 20M & 66.37 & 5.97 \\ \hline
\end{tabular}
}
\end{table}
\begin{table*}[t]
\centering
\caption{The table shows the Accuracy/F1 with ELMo.}
\label{elmo-best}
\resizebox{0.62\textwidth}{!}{
\begin{tabular}{lcccccc}
\hline
\multicolumn{1}{c}{\multirow{2}{*}{\textbf{\textbf{SST-5} }}} & \multicolumn{2}{c}{\textbf{r=10}} & \multicolumn{2}{c}{\textbf{r=400}} & \multicolumn{2}{c}{\textbf{Best}} \\ \cline{2-7} 
& \textbf{Acc.} & \textbf{E(r)} & \textbf{Acc.} & \textbf{E(r)} & \textbf{Acc. (avg)} & \textbf{E(r) (avg)} \\ \hline
BCN & - & - & - & - & 53.7{$^\ddagger$} & - \\
BCN + ELMo & - & - & - & - & 54.5{$^\ddagger$} & - \\ \hline
Semi-NMF $\mathbf{W}_h$ & 50.18 & 0.29 & 53.93 & 0.21 & 54.16 (52.93) & 0.09 (0.17)\\
SVD $\mathbf{W}_h$ & 50.4 & 0.27 & 54.11 & 0.13 & 54.11 (52.84) & 0.12 (0.17)\\
Pruning $\mathbf{W}_h$ & \textbf{50.81} & \textbf{0.25} & \textbf{54.66} & \textbf{-0.03} & \textbf{54.88} \textbf{(53.59)} & \textbf{-0.07} \textbf{(0.06)} \\ \hline 
Semi-NMF $\mathbf{W}_i$ & 38.23 & 1.1 & 54.11 & 0.15 & 54.11 (50.56) & 0.12 (0.34)\\
SVD $\mathbf{W}_i$ & 40.58 & 0.94 & 54.34 & 0.05 & 54.38 (51.19) & 0.02 (0.26) \\
Pruning $\mathbf{W}_i$ & 34.57 & 1.35 & 54.61 & -0.01 & 54.66 (50.01) & -0.02 (0.33) \\ \hline
\end{tabular}
}
\resizebox{0.62\textwidth}{!}{
\begin{tabular}{lcccccc}
\hline
\multicolumn{1}{c}{\multirow{2}{*}{\textbf{ \textbf{SNLI}}}} & \multicolumn{2}{c}{\textbf{r=10}} & \multicolumn{2}{c}{\textbf{r=400}} & \multicolumn{2}{c}{\textbf{Best}} \\ \cline{2-7} 
\multicolumn{1}{c}{} & \textbf{Acc.} & \textbf{E(r)} & \textbf{Acc.} & \textbf{E(r)} & \textbf{Acc. (avg)} & \textbf{E(r) (avg)} \\ \hline
ESIM & - & - & - & - & 88.6 & - \\
ESIM + ELMo & - & - & - & - & 88.5{$^\ddagger$} & - \\ \hline
Semi-NMF $\mathbf{W}_h$ & 87.24 & 0.04 & 88.45 & 0.01 & 88.47 (88.18) & 0.003 (0.01) \\
SVD $\mathbf{W}_h$ & 87.27 & 0.04 & 88.46 & 0.005 & 88.46 (88.18) & 0.003 (0.01) \\
Pruning $\mathbf{W}_h$ & \textbf{87.51} & \textbf{0.03} & \textbf{88.53} & \textbf{-0.003} & \textbf{88.53} \textbf{(88.23)} & \textbf{-0.003} (\textbf{0.01})  \\ \hline
Semi-NMF $\mathbf{W}_i$ & 77.08 & 0.39 & 88.44 & 0.01 & 88.44 (86.59) & 0.01 (0.07) \\
SVD $\mathbf{W}_i$ & 78.15 & 0.35 & 88.48 & 0.002 & 88.48 (86.77) & 0.007 (0.06) \\
Pruning $\mathbf{W}_i$ & 73.67 & 0.5 & 88.48 & 0.005 & 88.5 (85.8) & 0.001 (0.09) \\ \hline
\end{tabular}
}
\resizebox{0.62\textwidth}{!}{
\begin{tabular}{lcccccc}
\hline
\multicolumn{1}{c}{\multirow{2}{*}{\textbf{\textbf{SQuAD}}}} & \multicolumn{2}{c}{\textbf{r=10}} & \multicolumn{2}{c}{\textbf{r=400}} & \multicolumn{2}{c}{\textbf{Best}} \\ \cline{2-7} 
\multicolumn{1}{c}{} & \textbf{F1} & \textbf{E(r)} & \textbf{F1} & \textbf{E(r)} & \textbf{F1 (avg)} & \textbf{E(r) (avg)} \\ \hline
BiDAF & - & - & - & - & 77.3{$^\ddagger$} & - \\
BiDAF + ELMo & - & - & - & - & 81.75{$^\ddagger$} & - \\ \hline
Semi-NMF $\mathbf{W}_h$ & 76.59 & \textbf{0.21} & 81.55 & 0.04 & 81.55 (80.32) & 0.03 (0.07)\\
SVD $\mathbf{W}_h$ & \textbf{76.72} & \textbf{0.21} & 81.62 & 0.027 & 81.62 (80.47) & 0.02 (0.06)\\
Pruning $\mathbf{W}_h$ & 52.02 & 0.49 & 81.73 & 0.006 & 81.65 \textbf{(80.6)} & 0.006 
\textbf{(0.05)}\\ \hline
Semi-NMF $\mathbf{W}_i$ & 60.69 & 0.88 & \textbf{81.78} & \textbf{-0.0003} & \textbf{81.78} (77.93) & \textbf{-0.0003} (0.17) \\
SVD $\mathbf{W}_i$ & 57.14 & 1.03 & \textbf{81.78} & \textbf{-0.0003} & \textbf{81.78} (77.69) & \textbf{-0.0003} (0.17)\\
Pruning $\mathbf{W}_i$ & 52.02 & 1.24 & 81.73 & 0.004 & 81.73 (76.06) & 0.004 (0.25)  \\ \hline
\end{tabular}
}
\end{table*}

\section{Evaluation}
We evaluate using five different publicly available datasets spanning two domains: 1) Perplexity in two different Language Modeling (LM) datasets, 2) Accuracy/F1 in three downstream NLP tasks that ELMo achieved the state-of-the-art single-model performance. We also report the number of parameters, efficiency $E(r)$ (ratio of loss in performance to parameters compression), and inference time~\footnote{Using an Intel(R) Xeon(R) CPU E5-2620 v4 @2.10GHz.} in test set. 

We benchmark the LM capability using Penn Treebank~\cite[PTB]{Marcus:1993:BLA:972470.972475} and WikiText-2~\cite[WT2]{merity2016pointer}. For the downstream NLP tasks, we evaluate our method in the Stanford Question Answering Dataset~\cite[SQuAD]{squad} the Stanford Natural Language Inference~\cite[SNLI]{snli:emnlp2015} corpus, and the Stanford Sentiment Treebank~\cite[SST-5]{sst5} dataset. 

For all datasets, we run experiments across different levels of low-rank approximation \textit{r} with Semi-NMF and SVD, averaged over 5 runs, and compare with Pruning with same compression ratio. We also compare the factorization efficiency when only one of $\mathbf{W}_i$ or $\mathbf{W}_h$ was factorized. This is done in order to see which recurrence type (additive or multiplicative) is more suitable for compression.

\subsection{Measure}
For evaluating the performance of the compression we define efficiency measure as:
\begin{equation}
    E(r) = \frac{R(M,M^r)}{R(P,P^r)} 
\end{equation}
where $M$ represent any evaluation metric (i.e. Accuracy, F1-score, Perplexity\footnote{Note that for Perplexity, we use $R(M^r,M)$ instead, because lower is better.}), $P$ represents the number of parameters\footnote{$P^r$ and $M^r$ are the parameter and the measure after semi-NMF of rank $r$}, and $R(a,b)=\frac{a-b}{a}$ where $a=max(a,b)$, i.e. the ration. This indicator shows the ratio of loss in performance versus the loss in number of parameter. Hence, an efficient compression holds a very small $E$ since the denominator, $P - P^r$, became large just when the number of parameter decreases, and the numerator, $M - M^r$, became small only if there is no loss in the considered measure. In some cases $E$ became negative if there is an improvement.

\begin{figure*}[t]
    \centering
    \includegraphics[width=\textwidth]{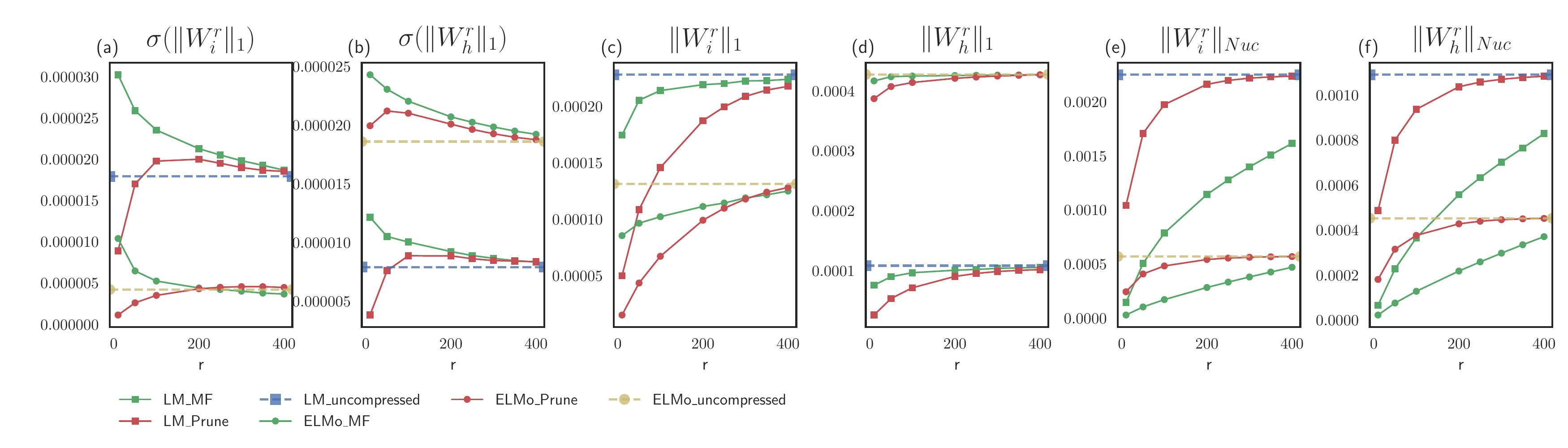}
    \caption{Norm analysis comparisons between MF and Pruning in Language Modeling (PTB) and ELMo.~Rank versus (a) $\sigma(\|\mathbf{W}_i\|_1)$ (b) $\sigma(\|\mathbf{W}_h\|_1)$ (c) $\|\mathbf{W}_i\|_{1}$ (d) $\|\mathbf{W}_h\|_{1}$ (e) $\|\mathbf{W}_i\|_{Nuc}$ (f)  $\|\mathbf{W}_h\|_{Nuc}$.  }
    \label{fig:Norm}
\end{figure*}

\subsection{Language Modeling (LM)}
We train a 3-layer LSTM Language Model proposed by~\cite{merity2018regularizing}, following the same training details for both datasets, using their released code~\footnote{https://github.com/salesforce/awd-lstm-lm}. In PTB, we fine-tune the compressed model for several epochs. Table \ref{lm} reports the perplexity among different ranks in $\mathbf{W}_{i,h}$. It is clear that compressing $\mathbf{W}_h$ works notably better than $\mathbf{W}_i$. We achieve similar results for WT-2. In general, SVD has the lowest perplexity among others. This difference becomes more evident for higher compression (e.g., r=10). Moreover, all the methods perform better than the result reported by~\cite{grachev2017neural} using Tensor Train (TT-LSTM). Using fine-tuning with rank 10 all the methods we achieve a small improvement compared to the baseline with a 2.13x speedup.

\subsection{NLP Tasks with ELMo}
To highlight the practicality of our proposed method, we also measure the factorization performances with models using pre-trained ELMo \cite{ELMo}, as ELMo is essentially a 2-layer bidirectional LSTM Language Model that captures rich contextualized representations. Using the same publicly released pre-trained ELMo weights~\footnote{https://allennlp.org/elmo} as the input embedding layer of all three tasks, we train publicly available state-of-the-art models as in \cite{ELMo}: BiDAF \cite{bidaf} for \textit{SQuAD}, ESIM \cite{chen2017enhanced} for \textit{SNLI}, and BCN \cite{mccann2017learned} for \textit{SST-5}. Similar to the Language Modeling tasks, we low-rank factorize the pre-trained ELMo layer only, and compare the accuracy and F1 scores across different levels of low-rank approximation. Note that although many of these models are based on RNNs, we factorize only the ELMo layer in order to show that our approach can effectively compress pre-trained transferable knowledge. As we only compress the ELMo weights, and other layers of each model also have large number of parameters, the inference time is affected less than in Language Modeling tasks. The percentage of parameters in the ELMo layer for BiDAF (\textit{SQuAD}) is 59.7\%, for ESIM (\textit{SNLI}) 67.4\%, and for BCN (\textit{SST-5}) 55.3\%.

\begin{figure}[!t]
    \centering
    \includegraphics[width=0.5\textwidth]{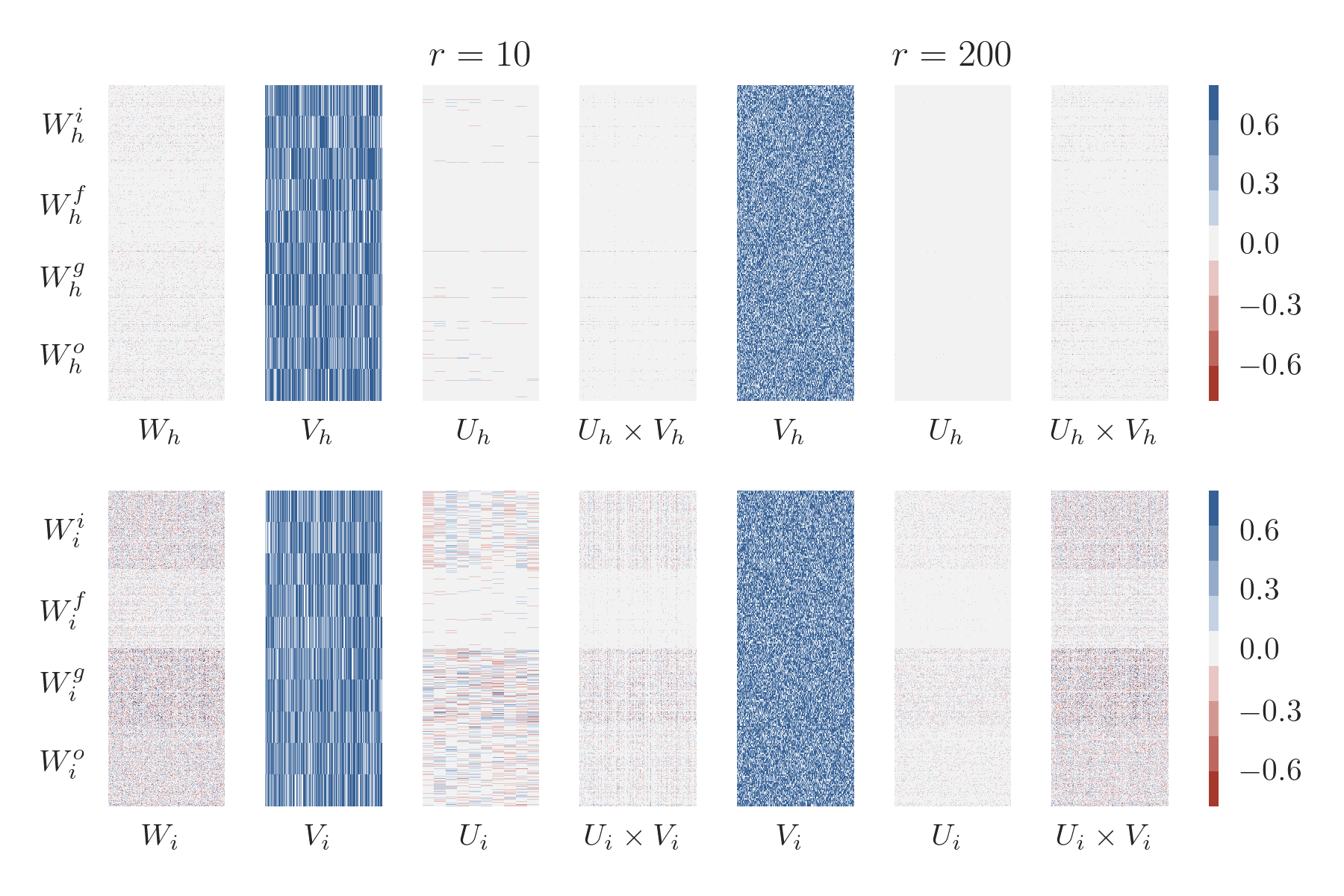}
    \caption{Heatmap LSTM weights on PTB.}
    \label{fig:heat}
\end{figure}

\begin{figure}[!t]
    \centering
    \includegraphics[width=0.5\textwidth]{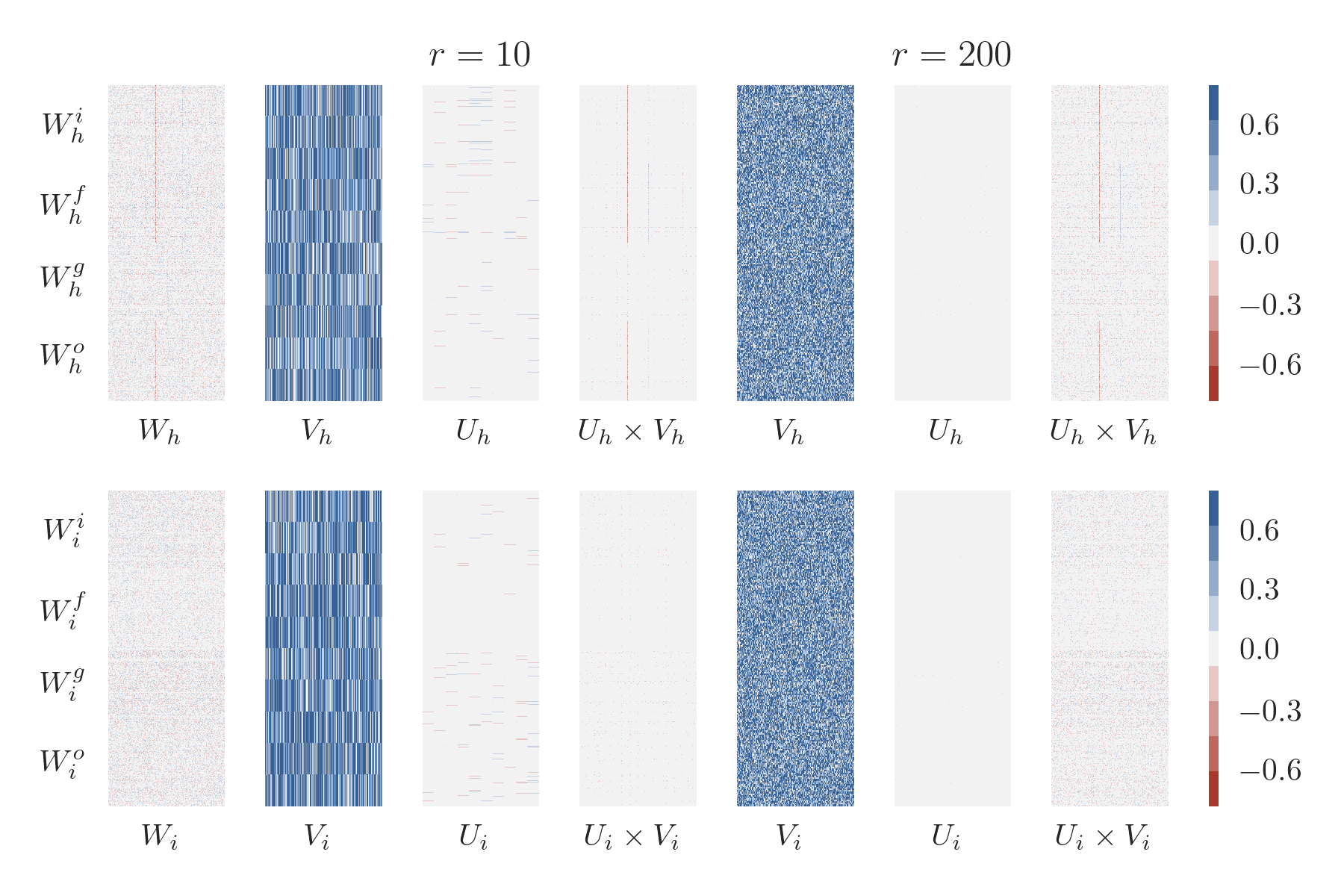}
    \caption{Heatmap of ELMo forward weights.}
    \label{fig:Normforwadwifht}
\end{figure}

From Table~\ref{elmo-best}, for \textit{SST-5} and \textit{SNLI}, we can see that compressing $\mathbf{W}_h$ is in general more efficient and better performing than compressing $\mathbf{W}_i$, except for SVD in \textit{SST-5}. On the other hand, for the results on SQuAD, Table~\ref{elmo-best} shows the opposite trend, in which compressing $\mathbf{W}_i$ constantly outperforms compressing $\mathbf{W}_h$ for all methods we experimented with. 
In fact, we can see that, in average, using highly compressed ELMo with BiDAF still performs better than without. Overall, we can see that for all datasets, we achieve performances that are not significantly different from the baseline results even after compressing over more than 10M parameters.

\subsection{Norm Analysis}
In the previous section, we observe two interesting points: 1) Matrix Factorization (MF) works consistently better in PTB and Wiki-Text 2, but Pruning works better in ELMo for $\mathbf{W}_h$, 2) Factorizing $\mathbf{W}_h$ is generally better than factorizing $\mathbf{W}_i$. To answer these questions, we collect the L1 norm and Nuclear norm statistics, defined in Figure \ref{fig:Norm}, comparing among $\mathbf{W}_h$ and $\mathbf{W}_i$ for both PTB and ELMo. L1 and its standard deviation (\textit{std}) together describe the sparsity of a matrix, and Nuclear norm approximates the matrix rank.

\paragraph{MF versus Pruning in $\mathbf{W}_i$}
From the results, we observe that MF performs better than Pruning in compressing $\mathbf{W}_i$ for high compression ratios. Figure~\ref{fig:Norm} shows rank $r$ versus L1 norm and its standard deviation, in both PTB and ELMo.
The first notable pattern from Figure~\ref{fig:Norm} Panel (a) is that MF and Pruning have diverging values from $r\leq200$. We can see that Pruning makes the \textit{std} of L1 lower than the uncompressed, while MF monotonically increases the \textit{std} from uncompressed baseline. This means that as we approximate to lower ranks ($r\leq200$), MF retains more salient information, while Pruning loses some of that salient information. This can be clearly shown from Panel (c), in which Pruning always drops significantly more in L1 than MF does. 

\paragraph{MF versus Pruning in $\mathbf{W}_h$}
The results for $\mathbf{W}_h$ are also consistent in both PTB and WT2; MF works better than Pruning for higher compression ratios. On the other hand, results from Table \ref{elmo-best} show that Pruning works better than MF in $\mathbf{W}_h$ of ELMo even in higher compression ratios. 

We can see from Panel (d) that L1 norms of MF and Pruning do not significantly deviate nor decrease much from the uncompressed baseline. Meanwhile, Panel (b) reveals an interesting pattern, in which the \textit{std} actually increases for Pruning and is always kept above the uncompressed baseline. This means that Pruning retains salient information for $\mathbf{W}_h$, while keeping the matrix sparse.

This behavior of $\mathbf{W}_h$ can be explained by the nature of the compression and with inherent matrix sparsity. In this setting, pruning is zeroing values already close to zero, so it is able to keep the L1 stable while increasing the \textit{std}. On the other hand, MF instead reduces noise by pushing lower values to be even lower (or zero) and keeps salient information by pushing larger values to be even larger. This pattern is more evident in Figure~\ref{fig:heat} and Figure~\ref{fig:Normforwadwifht}, in which you can see a clear salient red line in $\mathbf{W}_h$ that gets even stronger after factorization ($\mathbf{U}_h \times \mathbf{V}_h$). Naturally, when the compression rate is low (e.g., r=300) pruning is more efficient strategy then MF. 

\paragraph{$\mathbf{W}_i$ versus $\mathbf{W}_h$}
We show the change in Nuclear norm and their corresponding starting points (i.e., uncompressed) in Figure~\ref{fig:Norm} Panels (e) and (f). Notably, $\mathbf{W}_h$ has a consistently lower nuclear norm in both tasks compared to $\mathbf{W}_i$. This difference is larger for LM (PTB), in which $\|\mathbf{W}_i\|_{Nuc}$ is twice of that of $\|\mathbf{W}_h\|_{Nuc}$. By definition, having a lower nuclear norm is often an indicator of low-rank in a matrix; hence, we hypothesize that $\mathbf{W}_h$ is inherently low-rank than $\mathbf{W}_i$. We confirm this from Panel (d), in which even with a very high compression ratio (e.g., $r=10$), the L1 norm does not decrease that much. This explains the large gap in performance between the compression of $\mathbf{W}_i$ and $\mathbf{W}_h$. On the other hand, in ELMo, this gap in norm is lower and also shows smaller differences in performance between $\mathbf{W}_i$ and $\mathbf{W}_h$, and also sometimes even the opposite in SQuAD. Hence, we believe that smaller nuclear norms lead to better performance for all compression methods.

\section{Conclusion}
In conclusion, we empirically verified the limits of compressing LSTM gates using low-rank matrix factorization and pruning in four different NLP tasks. Our experiment results and norm analysis show that Low-Rank Matrix Factorization works better in general than pruning, except for particularly sparse matrices. We also discover that inherent low-rankness and low nuclear norm correlate well, explaining why compressing multiplicative recurrence works better than compressing additive recurrence. In future works, we plan to factorize all LSTMs in the model, e.g. BiDAF model, and try to combine both Pruning and Matrix Factorization.

\bibliography{acl2019}
\bibliographystyle{acl}

\end{document}